\documentclass[wcp]{jmlr}

\usepackage{longtable}% for long tables

\usepackage{booktabs}
\usepackage{multirow}
\usepackage{graphicx}

\usepackage{footmisc}

\makeatletter
\let\Ginclude@graphics\@org@Ginclude@graphics 
\makeatother

\jmlrvolume{}
\jmlryear{2023}
\jmlrworkshop{ACML 2023}

\title{ProtoDiffusion: Classifier-Free Diffusion Guidance with Prototype Learning}

\author{\Name{Gulcin Baykal}\thanks{Equal contribution.} 
\Email{baykalg@itu.edu.tr} \AND \Name{Halil Faruk Karagoz}\footnotemark[1] \Email{halilfarukkaragoz@gmail.com}
\\
\addr Istanbul Technical University, Computer Engineering Department, Istanbul, Turkey \AND
\Name{Taha Binhuraib} \Email{taha.huraib@gmail.com}\\
\addr Novus Technologies, Boston, USA \AND
\Name{Gozde Unal} \Email{gozde.unal@itu.edu.tr}\\
\addr Istanbul Technical University, AI and Data Engineering Department, Istanbul, Turkey}

\editors{Berrin Yan{\i}ko\u{g}lu and Wray Buntine}

\begin{document}

\maketitle

\begin{abstract}

Diffusion models are generative models that have shown significant advantages compared to other generative models in terms of higher generation quality and more stable training. However, the computational need for training diffusion models is considerably increased. In this work, we incorporate prototype learning into diffusion models to achieve high generation quality faster than the original diffusion model. Instead of randomly initialized class embeddings, we use separately learned class prototypes as the conditioning information to guide the diffusion process. We observe that our method, called ProtoDiffusion, achieves better performance in the early stages of training compared to the baseline method, signifying that using the learned prototypes shortens the training time. We demonstrate the performance of ProtoDiffusion using various datasets and experimental settings, achieving the best performance in shorter times across all settings.

\end{abstract}

\section{Introduction} \label{sec:intro}

Diffusion models have gained significant popularity for various image generation tasks due to their visually pleasing and perceptually accurate results. This popularity has especially increased after the Denoising Diffusion Probabilistic Models (DDPMs) are shown to outperform the state-of-the-art GAN-based methods for image synthesis (\cite{dhariwal2021diffusion}). Diffusion models are used for different tasks such as unconditional (\cite{ho2020denoising, song2021denoising}) and conditional (\cite{DBLP:conf/icml/NicholDRSMMSC22, rombach2021highresolutionis}) image synthesis, super-resolution (\cite{ho2021cascaded}), image inpainting (\cite{DBLP:conf/cvpr/LugmayrDRYTG22}), image-to-image translation (\cite{10.1145/3528233.3530757}), text-to-image synthesis (\cite{ramesh2022hierarchical, rombach2021highresolutionis}), and video synthesis (\cite{ho2022video}). While the diffusion models attain outstanding performance for these tasks, the reasons behind the success of the diffusion models as well as their existing problems are currently important venues of exploration.

Most of the aforementioned works utilize enriched conditioning information, such as text or images, and their high performance can mainly be attributed to the conditioning processes. While the conditioning enables high generation performance that is attained in relatively few sampling steps, handling text-based data is difficult. Therefore, diffusion models using ``lighter" conditioning constraints such as class labels are progressively being developed pertaining to their theoretical and practical aspects. Class labels, which can be considered as less enriched conditioning information, are used to guide the generation process in various ways (\cite{dhariwal2021diffusion, ho2021classifierfree}). However, obtaining stable and fast training of diffusion models using the class labels remains a significant challenge (\cite{DBLP:conf/icml/NicholDRSMMSC22}).

To mitigate the problem of fast training of the diffusion models, various solutions for decreasing the training time might be used. Performing parallelized training across multiple GPUs is one way to decrease the time, but it is computationally expensive. Additionally, novel diffusion models for increased stability using a few sampling steps might be designed as in (\cite{salimans2022progressive}). In our work, we search the ways of performing fast and stable training of the diffusion models that use the class labels as the conditioning information, without forming a new training method as in (\cite{salimans2022progressive}).

Classifier-free diffusion models (\cite{ho2021classifierfree}) use class labels as conditioning to guide the generation, and they do not need any explicit classifier for the training as in (\cite{dhariwal2021diffusion}). For each class in the dataset, the embeddings which represent these classes are learned during the training of the classifier-free diffusion models. These embeddings can be viewed as prototypes, and it is important to learn them effectively such that they can guide the generations with stability and consistency. As prototype learning by itself is a challenging task, we hypothesize that learning the prototypes during the diffusion process disregarding the prototype learning concepts might have significant effects on the consistency of the generations and the speed of the diffusion training. Therefore, exploring the prototype learning methods is crucially the first step to test this hypothesis.

Prototype learning is a favoured machine learning method for learning representative low-dimensional embeddings to represent the diverse classes in the data. More than one prototype might be used to represent a single class to capture the essential characteristics of that class. Generally, they are updated based on specifically designed objectives during the training of the learning model. For the classification of the data, the distances to the learned prototypes are used to assign the data to the closest prototype's class. The effectiveness of the prototype learning is presented for various tasks (\cite{DBLP:conf/cvpr/YangZYL18, DBLP:conf/cvpr/ZhouWKG22, Shu2019PODNPO, Kim2021PrototypeGuidedSF, xu2020attribute}), and we find out that carefully designed methods for learning the prototypes actually matter in terms of performance. Therefore, we study on how the prototype learning can be incorporated within the diffusion models to mitigate the stabilized and fast training problems.

In this work, we propose a novel combination of prototype learning and diffusion models called \textit{ProtoDiffusion}. We learn the prototypes of the classes with a separate classifier (\cite{DBLP:conf/cvpr/YangZYL18}) in $\sim 2-10$ minutes, based on the dataset. Then, we start the training of the diffusion model after we initialize the class embeddings with the learned prototypes, which are used to guide the diffusion process. We demonstrate that the quality of the generated images by the diffusion model is already increased by the beginning of the training. Therefore, our model achieves the same performance $\sim 2$ times faster than the original diffusion model which indicates a significant amount of reduction in the training time. Hence, training a model to learn the prototypes in at most 10 minutes saves hours in the training of the diffusion model. We summarize our contributions as follows:

\begin{itemize}
    \item We present that the representativeness of the prototypes used as the conditioning information for the diffusion models might affect the training time and the generation performance.
    \item We mitigate the training time problem of the diffusion models using the already learned prototypes of the classes.
    \item We demonstrate that the stability of the training in terms of generation quality is increased, and this stability even leads to better generation quality compared to the baseline.
\end{itemize}

The ProtoDiffusion model we present in this work outperforms the performance of the baseline methods in terms of Fr$\acute{e}$chet Inception Distance (FID) (\cite{heusel2017gans}) and Inception Score (IS) (\cite{salimans2016improved}) on various datasets while we obtain those results faster than the baselines.

\section{Related Work} \label{sec:rel_work}

\paragraph{Diffusion Models} Diffusion models are promising generative models that have the potential to generate high-quality images with a variety of applications. While some of the works focus on the decreased sampling time (\cite{song2021denoising, salimans2022progressive, zhang2023fast}), some other works propose architectural improvements to increase the generation quality such as cascaded diffusion (\cite{ho2021cascaded}). Alongside with these methods, conditioning the diffusion process with enriched information such as text and image to improve the generation quality is a highly studied approach. Generally, the text and the image information are embedded to the latent spaces using pretrained encoders such as CLIP (\cite{radford21clip}) or VQGAN (\cite{Esser2020TamingTF}) to ease the training of the enormous diffusion models as in DALL-E (\cite{ramesh2021zeroshot}), DALL-E 2 (\cite{ramesh2022hierarchical}), and Imagen (\cite{saharia2022photorealistic}). Further models utilizing the conditioning information like Latent Diffusion Models (LDMs) (\cite{rombach2021highresolutionis}) perform the diffusion process over the latent representation instead of the image in order to decrease the need for computational power. Retrieval augmented diffusion models (\cite{blattmann22retrieval}) condition the image generation on the neighbors of the input image. The neighbor images are retrieved from a database, and they enable achieving a better performance compared to the baseline using a comparably smaller diffusion model. Even though our method employs simpler information as the class label for conditioning, it has a similar intuition to the retrieval augmented diffusion models. As we incorporate the learned prototypes of the classes for conditioning, each of these prototypes captures the various aspects of the class that it represents. Therefore, the learned prototypes have the similar advantage of using the retrieved neighbors which also capture the different aspects of the class that the input image belongs to.

\paragraph{Prototype Learning} Prototype learning methods utilizing deep neural networks have gained interest for a wide range of tasks. \cite{DBLP:conf/cvpr/ZhouWKG22} propose a semantic segmentation solution with a prototype view. They learn to represent different semantic classes with prototypes to improve their performance. \cite{xu2020attribute} employ prototype learning to zero-shot representation learning while \cite{snell17proto} propose prototypical networks for few-shot classification problem. \cite{Shu2019PODNPO} propose a prototype-based network for open set recognition tasks. \cite{Kim2021PrototypeGuidedSF} utilize prototypes for saliency feature learning for person search. \cite{oord2017neural} use vector quantization to learn a discrete latent space for image reconstruction task that the discrete latent space essentially includes prototypes. Unsupervised (\cite{Wu2018UnsupervisedFL}) and supervised (\cite{GuerrieroICLR2018, mettes19hyperspherical, DBLP:conf/cvpr/YangZYL18}) classification can be also performed using prototypes. In our work, we need to learn prototypes to effectively represent each of the classes in the dataset. As we have a supervised setting, we use the method proposed by \cite{DBLP:conf/cvpr/YangZYL18} which performs supervised classification by learning prototypes. After learning such prototypes, we use them to guide the diffusion training.

\section{Background} \label{sec:background}

\subsection{Diffusion Models} \label{sec:background_diffusion}

We use Gaussian diffusion models as a generative method. After the diffusion models are introduced by \cite{pmlr-v37-sohl-dickstein15}, numerous models based on the diffusion idea are proposed. In the training phase of the diffusion models, Gaussian noise is progressively added to a sample from the real data distribution $x_0 \sim q(x)$ in $T$ time steps:

\begin{equation}
    q(x_t|x_{t-1}) = \mathcal{N}(x_t; \sqrt{1-\beta_t}x_{t-1}, \beta_t\mathrm{I}).
    \label{eq: noising}
\end{equation}
where $x_{t-1}$ is transformed to $x_t$ by adding Gaussian noise at $t^{th}$ time step. $\beta_t$ is the variance of the added noise, and $\sqrt{1-\beta_t}$ is the scaling parameter according to a variance schedule. In order to obtain $x_t$ at an arbitrary time step $t$ without $t$ steps iterations, a reparameterization trick enabled by the properties of the Gaussian distribution might be performed:

\begin{align}
    q(x_t|x_0) &= \mathcal{N}(x_t; \sqrt{\bar{\alpha_t}}x_0, (1-\bar{\alpha}_t)\mathrm{I}), \nonumber \\
    x_t &= \sqrt{\bar{\alpha}_t}x_0 + \sqrt{1-\bar{\alpha}_t}\epsilon. 
    \label{eq: reparam}
\end{align}
where $\alpha_t = 1- \beta_t$, $\bar{\alpha}_t = \prod_{i=1}^t \alpha_i$, and $\epsilon \sim \mathcal{N}(0, \mathrm{I})$.

The training phase of the diffusion models consists of the denoising process to reverse the noising process in Equation~\ref{eq: noising}. The denoising process is defined as:

\begin{equation}
    p_\theta(x_{t-1}|x_t) = \mathcal{N}(x_{t-1}; \mu_\theta(x_t, t), \Sigma_\theta(x_t, t)).
    \label{eq: denoising}
\end{equation}
where $\theta$ defines the parameters of the neural network that predicts the mean $\mu_\theta(x_t, t)$ and the variance $\Sigma_\theta(x_t, t)$ parameters of the Gaussian distribution. Starting from the white Gaussian noise $x_T$, the $x_0$ image is obtained by gradually reducing the noise in $T$ time steps. As a generative model, the main objective of the diffusion models is to learn the reverse process such that a high quality image $x_0$ can be attained using a random noise vector $x_T$. In DDPM (\cite{ho2020denoising}), the mean $\mu_\theta(x_t, t)$ is learned while the variance $\Sigma_\theta(x_t, t)$ is set to constant.

A tractable variational lower bound exists for the optimization of the neural network in Equation~\ref{eq: denoising}. \cite{ho2020denoising} decompose the objective function, and presents that predicting the noise $\epsilon$ added in the current time step as in Equation~\ref{eq: reparam} is the best way to parameterize the mean $\mu_\theta(x_t, t)$ of the model:

\begin{equation}
    \mu_\theta(x_t, t) = \frac{1}{\sqrt{\alpha_t}}\left ( x_t - \frac{\beta_t}{\sqrt{1-\bar{\alpha}_t}}\epsilon_\theta(x_t, t)\right ).
\end{equation}
where $\epsilon_\theta$ is the function approximator to predict the $\epsilon$ value added to obtain $x_t$. The simplified training objective is derived as in (\cite{ho2020denoising}):

\begin{equation}
    L_{simple} = \mathbb{E}_{t \sim [1, T], x_0 \sim q(x), \epsilon \sim \mathcal{N}(0, \mathrm{I})}[||\epsilon - \epsilon_\theta(x_t, t)||^2].
    \label{eq: ddpm_loss}
\end{equation}

While the DDPM synthesizes unconditional images, guided diffusion models are also available for conditional image generation. \cite{dhariwal2021diffusion} propose classifier guidance where the class conditional parameters $\mu_\theta(x_t|y)$ and $\Sigma_\theta(x_t|y)$ are perturbed by the gradients of a classifier $p_\phi(y|x_t)$ that predicts the target class $y$. The perturbed mean with the guidance scale $s$ is derived as:

\begin{equation}
    \title{\mu}_\theta(x_t|y) = \mu_\theta(x_t|y) + s\Sigma_\theta(x_t|y)\nabla_{x_t}\text{log}(p_\phi(y|x_t)).
    \label{eq: classifier_guidance}
\end{equation}

Even though the classifier guidance improves the quality of the images, there are a few problems of the classifier guidance. As the denoising process starts with highly noised input and proceeds with noisy images at most of the time steps, the classifier should be robust to noise. While obtaining such a classifier is challenging, predicting a class label does not require using the most of the information in the data. Therefore, taking the gradients of such a classifier might mislead the generation direction. 

\cite{ho2021classifierfree} propose classifier-free guidance method that does not require a separate classifier. The conditioning information $y$ is periodically utilized while it is dropped out at the remaining time. Therefore, a single model can be used for both unconditional and conditional generation. \cite{ho2021classifierfree} derive that the unconditional $\epsilon_\theta(x_t, t)$ and conditional $\epsilon_\theta(x_t, t, y)$ estimations can be used to represent the gradients of the classifier as:

\begin{align}
    \nabla_{x_t}\text{log}(y|x_t) &= \nabla_{x_t}\text{log}p(x_t|y) - \nabla_{x_t}\text{log}p(x_t) \nonumber \\
    &= -\frac{1}{\sqrt{1-\bar{\alpha}_t}}\left( \epsilon_\theta(x_t, t, y) - \epsilon_\theta(x_t, t) \right).
    \label{eq: classifier_free}
\end{align}

Equation~\ref{eq: classifier_free} implies that an implicit classifier can eliminate the need for an explicit classifier, and \cite{ho2021classifierfree} report better results with the classifier-free guidance compared to the explicit classifier guidance. 

In our work, we enhance the classifier-free guidance method with prototype learning in order to improve the performance while decreasing the training time.

\subsection{Prototype Learning} \label{sec:background_proto_learning}

Prototype learning is a classical and popular representation learning method for numerous tasks as described in Section~\ref{sec:rel_work}. Specifically, prototype learning is a plausible choice to categorize the labeled data or to cluster the unlabeled data. For a classification task, prototypes that define different classes might be learned through a parameter optimization of feature extractor convolutional neural networks (CNNs). The extracted features by the CNN might be compared to the prototypes of the classes to predict the target class instead of using softmax. The CNN parameters and the prototypes might be jointly trained. 

Given $C$ different classes in the dataset and $K$ different prototypes for each class, a prototype is denoted as $e_{ij}$ where $i \in {1, 2, \ldots, C}$ and $j \in {1, 2, \ldots, K}$. The number of prototypes for each class $K$ is a design choice. The input data $x$ is given to a CNN $f_\psi$ for feature extraction, and the features of $x$ might be compared to the prototypes for the class assignment:

\begin{equation}
    x \in \text{argmax}_{i=1}^C (-\text{min}_{j=1}^K ||f_\psi(x)-e_{ij}||_2^2).
    \label{eq: feedforward}
\end{equation}

Via Equation~\ref{eq: feedforward}, the input is assigned to the class which consists of the closest prototype to the input's features. In order to jointly train the feature extractor $f_\psi$ and the codebook $\mathcal{M}$, which comprises the prototypes $e$'s, in an end-to-end manner, carefully designed loss functions might be used. \cite{DBLP:conf/cvpr/YangZYL18} introduce several loss functions for the training, and compares their effects on the accuracy of the classification. The withstanding loss function is defined as:

\begin{align}
    p(x \in e_{ij} | x) &= \frac{e^{-\gamma||f_\psi(x)-e_{ij}||_2^2}}{\sum_{k=1}^C\sum_{l=1}^K e^{-\gamma||f_\psi(x)-e_{kl}||_2^2}}, \label{eq: prob} \\
    \mathcal{L}(\psi, \mathcal{M}) &= -\text{log}\left(\sum_{j=1}^K p(x \in e_{yj} | x)\right) + \lambda ||f_\psi(x)-e_{\text{closest}}||_2^2. \label{eq: classification_loss}
\end{align}

The first term in Equation~\ref{eq: classification_loss} minimizes the cross entropy loss using a sample $(x, y)$ and its probability of belonging to the prototype $e_{ij}$ defined in Equation~\ref{eq: prob}. In Equation~\ref{eq: prob}, the $\gamma$ parameter controls the sharpness of the assignment, and the probability of belonging to a prototype is calculated using the distance to the features extracted by $f_\psi$. 

While minimizing the first term in the loss function decreases the sample's distance to the closest prototype compared to the distances to the remaining prototypes, it is not enough to minimize the distance between the sample and the closest prototype. Therefore, a regularization term forcing the minimum distance between the sample and the closest prototype $e_\text{closest}$ is added to the optimization as in the second term in Equation~\ref{eq: classification_loss}. Training $f_\psi$ and $\mathcal{M}$ using this loss function leads to better classification results on numerous datasets as presented in (\cite{DBLP:conf/cvpr/YangZYL18}).

In our work, we exploit the learned prototypes that represent the classes in the data space in order to enhance the classifier-free diffusion guidance.

\section{Method} \label{sec:method}

\begin{figure}[t!]
    \centering
    \includegraphics[width=\textwidth]{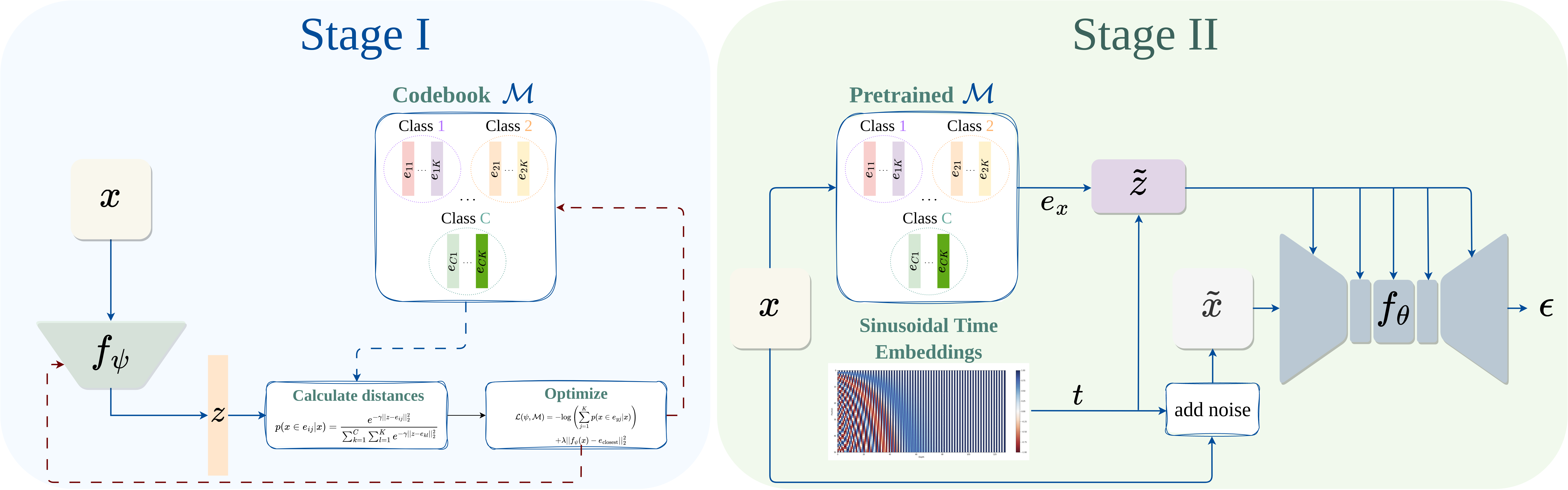}
    \caption{Training stages of ProtoDiffusion. At the first stage, a codebook $\mathcal{M}$ consisting of the prototypes are trained. Then, the pretrained prototypes and the time embeddings are concatenated as $\tilde{z}$ to guide the diffusion process based on the class and the time information. Lastly, the noisy image $\tilde{x}$ is forwarded to a U-Net like network for the noise prediction.}
    \label{fig:main_figure}
\end{figure}

Our model ProtoDiffusion is presented in Figure~\ref{fig:main_figure}. We conduct two staged training such that at the first stage, we obtain the class based prototypes while at the second stage, we train the diffusion model.

At Stage I, we follow the prototype learning method (\cite{DBLP:conf/cvpr/YangZYL18}) explained in Section~\ref{sec:background_proto_learning}. Based on the dataset, the number of the classes in the codebook $\mathcal{M}$ may vary. However, to be able to use the class embeddings in the diffusion process as in (\cite{ho2021classifierfree}), we need a single embedding to represent the class. Therefore, we learn a single prototype per class in Stage I which is practically allowed as shown in (\cite{DBLP:conf/cvpr/YangZYL18}). As illustrated in Stage I of Figure~\ref{fig:main_figure}, we train the feature extractor $f_\psi$ and the codebook $\mathcal{M}$ in an end-to-end manner.

At Stage II, we follow the proposed diffusion training proposed by \cite{ho2021classifierfree} with specific differences as follows. Instead of initializing the class embeddings randomly, we utilize the prototypes that we learned in Stage I, and initialize the class embeddings with our prototypes. Then, we concatenate the sampled sinusoidal time embedding $t$ with the prototype $e_{x}$ that represents the class label of the input $x$, and obtain $\tilde{z}$. As explained in Section~\ref{sec:background_diffusion}, the denoising process is learned by a neural network, and it is optimized by predicting the added noise to the input $x$ in $t$ time steps. Stage II in Figure~\ref{fig:main_figure} consists of a U-Net like neural network $f_\theta$ to predict the noise $\epsilon$ of the noisy image $\tilde{x}$. During the training, various time steps are sampled in different iterations, and $f_\theta$ is optimized to predict the noise values based on these time steps. In order to guide the diffusion process, the class embedding and the time step are also given to the specific layers of $f_\theta$. 

In the baseline model, the randomly initialized class embeddings are optimized by the diffusion process without explicit optimizations measuring the representativeness of these embeddings. Therefore, we suspect that attaining prototypical class embeddings with this setting might be a bottleneck for the training time and the generation quality. As a result, we start the diffusion training with already representative class embeddings, and achieve a better performance in a shorter time period. We incorporate the learned prototypes to the diffusion process in two ways.

The first method we use is the ProtoDiffusion that utilizes the class prototypes as they are, and not update them during the diffusion training. We call it ProtoDiffusion (Frozen). The reason we follow this method is that the class prototypes are already learned, and they may not need further optimization to guide the diffusion training. Therefore, the gradients are not allowed to flow to the class embeddings to update them.

The second method we use is the ProtoDiffusion (Unfrozen) where the class prototypes are also updated during the diffusion training. Comparing these two methods with the baseline method demonstrates not only the effects of the prototypes to the diffusion training, but also the effects of the diffusion training to the representativeness of the class embeddings.  
\section{Experiments} \label{sec:experiments}

We conduct our experiments on CIFAR10 (\cite{Krizhevsky09}), STL10 (\cite{stl10}), and Tiny ImageNet (\cite{Le2015TinyIV}) datasets using 4 NVIDIA A100 GPUs. We present all of the experimental settings in Appendix~\ref{apd:experimental_setting}. 

We compare the GPU hours of the baseline diffusion model and our model ProtoDiffusion until they reach their best performance in Table~\ref{tab:gpu_hours}. The results for the ProtoDiffusion models consist of the prototype learning time and the diffusion training time. We learn the prototypes for the datasets using a single NVIDIA A100 GPU. We obtain the prototypes of CIFAR10 and STL10 in $\sim 2$ minutes and Tiny ImageNet in $\sim 13$ minutes. ProtoDiffusion achieves the best performance considerably faster than the baseline model for all datasets, which supports our hypothesis for faster training enabled by the learned prototypes.

\begin{table}
\caption{GPU hours comparison of the baseline model and ProtoDiffusion model across datasets.}
\label{tab:gpu_hours}
\centering
\begin{tabular}{|c|c|c|c|}
    \hline
    {\textbf{Method}} &{\textbf{CIFAR10}} &{\textbf{STL10}} &{\textbf{Tiny ImageNet}}\\
    \hline
    Classifier-Free Guidance & 29.33 & 88.89 & 233.33 \\
    ProtoDiffusion & \textbf{13.33} & \textbf{66.67} & \textbf{166.67} \\
    \hline
\end{tabular}
\end{table}

\begin{figure}
    \centering
    \includegraphics[width=0.7\textwidth]{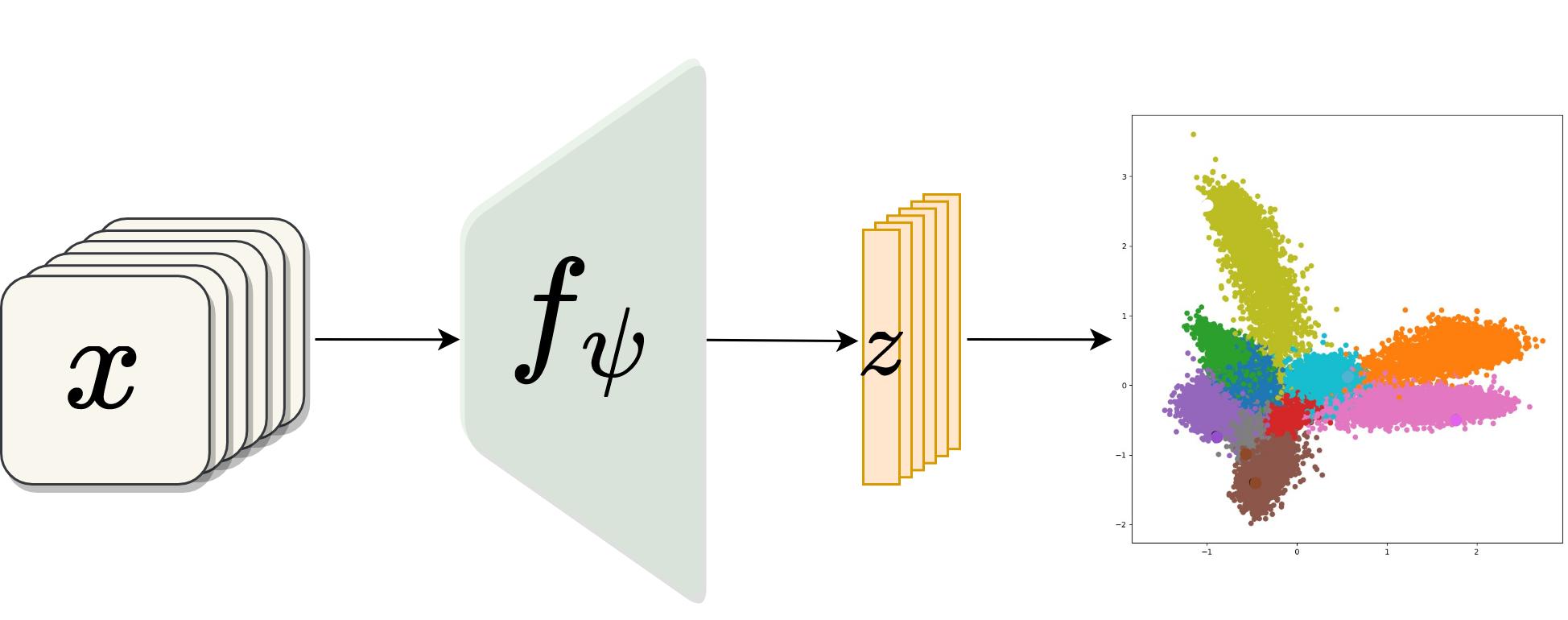}
    \caption{CIFAR10 clusters are depicted on the right after the prototype learning process.}
    \label{fig:cifar_clusters}
\end{figure}

We use an existing method for prototype learning proposed by \cite{DBLP:conf/cvpr/YangZYL18}. In order to  evaluate the performance of the learned feature extractor $f_\psi$, we project the data of the CIFAR10 dataset to a 2D space using PCA. As Figure~\ref{fig:cifar_clusters} shows, $f_\psi$ learns to cluster the data $x$ into classes accurately, and the learned prototypes in ProtoDiffusion are the cluster centers of these clusters.

We evaluate the performance of the baseline method and the ProtoDiffusion methods using FID (\cite{heusel2017gans}) and IS (\cite{salimans2016improved}) in Table~\ref{tab:eval_scores}. As the lower FID and higher IS indicate better performance, we show that ProtoDiffusion methods outperform the baseline method for all of the datasets, except for the IS measure with the Tiny ImageNet. We observe that the ProtoDiffusion (Unfrozen) method is mostly better than the ProtoDiffusion (Frozen) method. These results demonstrate that the parameter update of the class embeddings caused by the diffusion process enhances the stability and positively impacts training, indicating that the gradient updates from the diffusion process allows the embeddings to capture more of the latent space that represent the class labels. The FID, which compares the distribution of generated images with the distribution of real images in a given dataset, is a commonly preferred metric over the IS, which evaluates only the distribution of generated images. For the Tiny ImageNet dataset, although the IS scores of ProtoDiffusion models are below that of the baseline, both ProtoDiffusion models achieve lower FID scores than the baseline. 

\begin{table}
\caption{Comparison of the models in terms of FID ($\downarrow$) and IS ($\uparrow$) on different datasets.}
\label{tab:eval_scores}
\centering
\begin{tabular}{|c|cc|cc|cc|}
    \hline
    \multirow{2}{*}{\textbf{Method}} &\multicolumn{2}{c|}{\textbf{CIFAR10}} &\multicolumn{2}{c|}{\textbf{STL10}} &\multicolumn{2}{c|}{\textbf{Tiny ImageNet}}\\
    \cline{2-7}
    &\multicolumn{1}{c}{\textit{FID($\downarrow$)}} &\multicolumn{1}{c|}{\textit{IS($\uparrow$)}} &\multicolumn{1}{c}{\textit{FID($\downarrow$)}} &\multicolumn{1}{c|}{\textit{IS($\uparrow$)}} &\multicolumn{1}{c}{\textit{FID($\downarrow$)}} &\multicolumn{1}{c|}{\textit{IS($\uparrow$)}}\\
    \hline
    Classifier-Free Guidance & 11.82 & 8.85 & 33.86 & 11.33 & 11.49 & \textbf{28.67} \\
    ProtoDiffusion (Frozen) & 9.19 & 8.95 & 23.64 & 11.36 & \textbf{9.56} & 24.82 \\
    ProtoDiffusion (Unfrozen) & \textbf{8.55} & \textbf{9.01} & \textbf{22.57} & \textbf{11.51} & 9.94 & 23.07 \\
    \hline
\end{tabular}
\end{table}

\begin{figure}
    \centering
    \includegraphics[width=\textwidth]{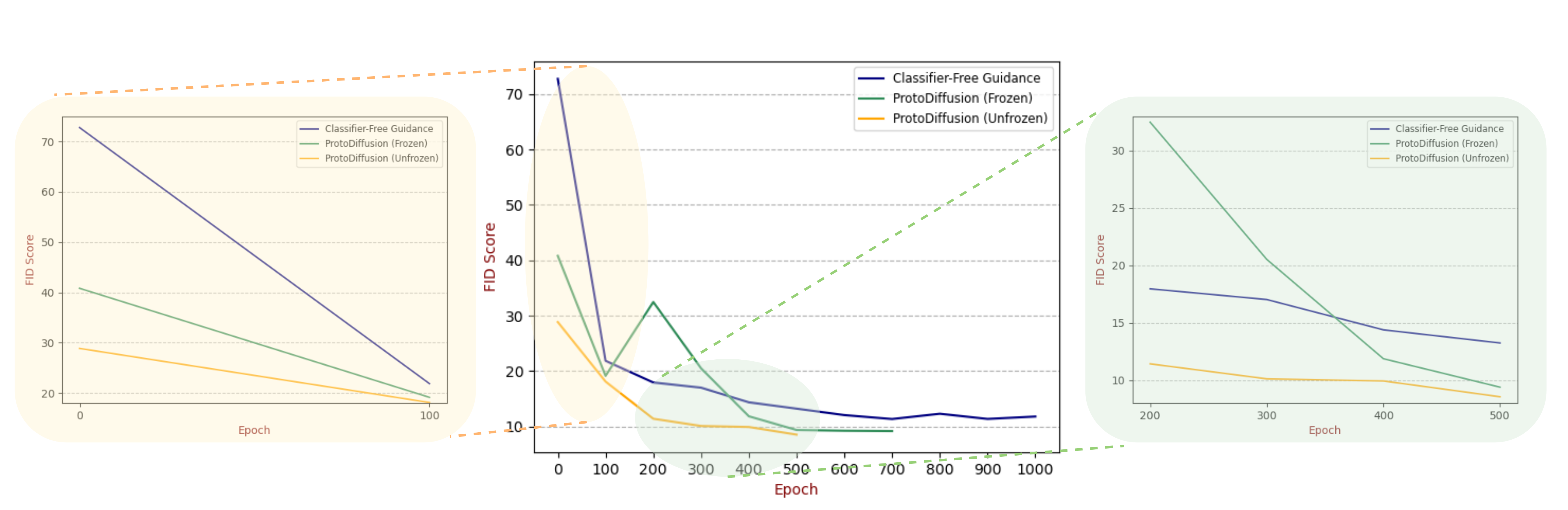}
    \caption{FID scores of all models during the training. While the yellow area highlights the benefits of using the learned prototypes to the starting, the green area highlights that the best results obtained for the dataset is achieved way faster than the baseline.}
    \label{fig:fid_chart}
\end{figure}

We monitor the training progress of all models so that we can decide whether to stop the training or not. Figure~\ref{fig:fid_chart} presents the FID scores calculated during the training of the models using the CIFAR10 dataset. For each 100 epoch, we check the FID values and compare them. The first important observation that we note is the large difference between the FID values of the baseline and the ProtoDiffusion models at the beginning of the training. We observe that the learned prototypes' effect on the generation quality at the beginning is considered valuable. Therefore, we manage to obtain the best performance faster than the baseline model. The green area highlights the epoch that we decide to stop the training for the ProtoDiffusion (Unfrozen) model. We compare the FID scores at the same epoch, and obtain a better score compared to that of the baseline. Therefore, we stop the training of the ProtoDiffusion (Unfrozen) at $500^{th}$ epoch while we continue to train the other models. We demonstrate that the ProtoDiffusion (Frozen) does not present a stabilized training compared to the ProtoDiffusion (Unfrozen). These results justify our conclusion that the frozen prototypes may not be adapted to the diffusion training in the same time frame as the ProtoDiffusion (Unfrozen) method since they are not updated with the gradients of the diffusion process. As we attain the best performance two times faster than the baseline, this also affects the computational usage demonstrated in Table~\ref{tab:gpu_hours}.

As an ablation study, we observe the effects of the prototype dimensionality on the performance in terms of generating quality. Table~\ref{tab:cdim} presents the FID results on the CIFAR10 dataset using 32, 64, and 128 as the prototype dimensionalities. While ProtoDiffusion (Unfrozen) outperforms the other methods for all settings, using 128 as the prototype dimensionality gives the best performance. Since the prototypes with higher dimensionalities can accommodate extra information about the underlying data space, it is reasonable to obtain a better performance in the latter case. Therefore, we choose 128 among the other options for all experiments.

\begin{table}
\caption{The effects of the different prototype dimensionalities on CIFAR10 in terms of FID.}
\label{tab:cdim}
\centering
\begin{tabular}{|c|c|c|c|}
    \hline
    \textbf{Method} & \textbf{32} & \textbf{64} & \textbf{128} \\
    \hline
    Classifier-Free Guidance & 12.51 & 9.89 & 11.82 \\
    ProtoDiffusion (Frozen)  & 9.89 & 12.48 & 9.98 \\
    ProtoDiffusion (Unfrozen) & \textbf{9.25} & \textbf{9.15} & \textbf{8.55} \\
    \hline
\end{tabular}
\end{table}

\begin{figure}[t]
    \label{fig:stl_generation_results}
    \caption{Generated samples from the STL10 dataset using the baseline and ProtoDiffusion.}
    \subfigure[Regular Classifier-Free Diffusion Guidance.]{
        \centering
        \includegraphics[width=0.3\textwidth]{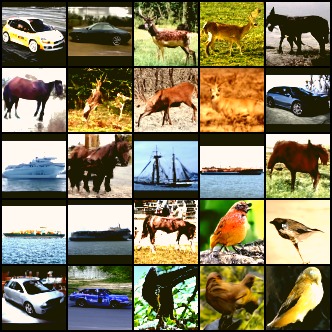}
        }
    \hfill
    \subfigure[ProtoDiffusion (Frozen).]{
        \centering
        \includegraphics[width=0.3\textwidth]{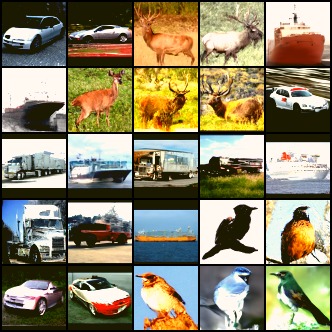}
        }
    \hfill
    \subfigure[ProtoDiffusion (Unfrozen).]{
        \centering
        \includegraphics[width=0.3\textwidth]{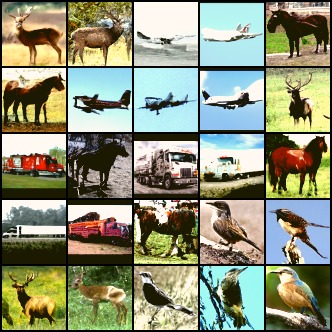}
        }
\end{figure}

In addition to the numerical evaluations, we perform visual assessments of the methods. Figure~\ref{fig:stl_generation_results} demonstrates the generated samples from the STL10 dataset using the baseline method and ProtoDiffusion methods. We observe that the generated samples by the ProtoDiffusion methods depict structural consistency compared to the baseline's samples. Additionally, the samples of almost all classes show good quality, while the generated samples from some classes such as the horse seem to appear more flawed.

\begin{figure}
    \centering
    \includegraphics[width=0.75\textwidth]{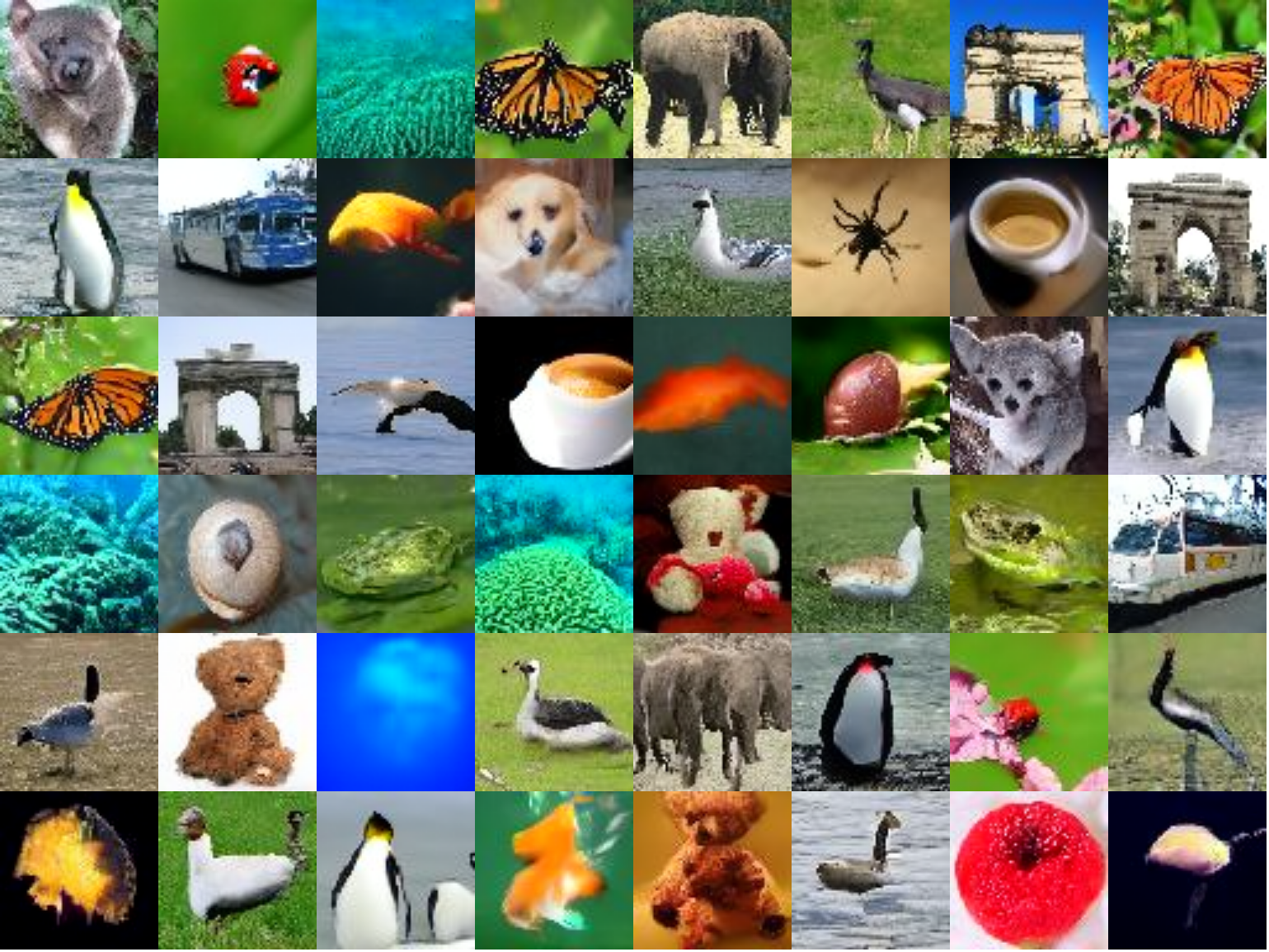}
    \caption{Generated samples from Tiny ImageNet dataset using ProtoDiffusion.}
    \label{fig:tiny_generation_results}
\end{figure}

We present generated samples from the Tiny ImageNet dataset using the best performing ProtoDiffusion method in Figure~\ref{fig:tiny_generation_results}. As the results depict, the ProtoDiffusion model produces samples from the diverse classes of the dataset showing good quality. 

\section{Conclusion} \label{sec:conclusion}

In this work, we propose a method called ProtoDiffusion that incorporates the representative class prototypes into the diffusion training as the conditioning information. We observe that the learned prototypes affect the training even at the beginning, and the improved performances in terms of generation quality are achieved faster than the baseline method. We conduct several experiments to visualize the representativeness of the prototypes, to view the effects of the prototype dimensionality, and to monitor the diffusion training progress so that we can derive conclusions about the significance of the prototype learning. 

Although the ProtoDiffusion models achieve significantly better performances way faster than the baseline for all datasets, we note that there is room for improvement for the Tiny ImageNet dataset. The effects of the prototype learning for the Tiny ImageNet dataset is not as clear as compared to the other datasets in terms of the chosen metrics. One reason might be the representativeness of the prototypes itself as they are attained by the same simple neural network which may not be adequate in  characterizing the complexity of the given dataset. Therefore, as future work, more complex datasets like Tiny ImageNet should be further studied to analyze the effects of the prototype learning.

%\acks{Acknowledgements should go at the end, before appendices and references. You can uncomment this for the camera-ready version on paper acceptance.}

%\bibliographystyle{plain}
\bibliography{acml23}

\appendix

\section{Experimental Settings}\label{apd:experimental_setting}

\subsection{Prototype Learning}

In the prototype learning, we use a basic ResNet-18 \cite{he2016residual} architecture as the feature extractor. We train the models 20 epochs for all datasets as we achieve the highest accuracy within the first 20 epochs. As presented in Table~\ref{tab:cdim}, we train different networks to learn prototypes with various dimensionalities. We use 512 batch size and Adam optimizer with 0.0001 learning rate.

\subsection{Diffusion Training}

In the diffusion training process, the U-Net architecture is utilized to effectively predict the noise present in Denoising Diffusion Probabilistic Models (DDPMs). The initial channel size of the model is set to 64. For the Encoder part, which is responsible for down-sampling the input, there are 4 blocks, each consisting of 2 ResNet layers. These layers are equipped with a 0.1 Dropout layer. The channel size of each block progressively increases using multiplication factors of 2, 2, 4, and 4. This hierarchical representation enables the capturing of both fine-grained and high-level features. The Decoder part, responsible for up-sampling, mirrors the Encoder by utilizing Upsample blocks with the same multiplication factors. However, the channel sizes of the Decoder's blocks gradually decrease, ultimately resulting in an RGB image with 3 channels. 

For all experiments, we adopt the AdamW optimizer with a fixed learning rate of 0.0002. The batch size of 128, which corresponds to one GPU, is used for CIFAR10 experiments. For the STL10 and Tiny ImageNet datasets, a batch size of 64 (one GPU) is employed.

In order to decrease the computational and the structural complexity during our experiments, we resize the image size of the STL10 dataset from 96 to 64. The original image sizes of the CIFAR10 and Tiny ImageNet datasets remain unchanged at 32 and 64, respectively.

\end{document}